%% file: UniV2X.tex
\documentclass[letterpaper]{article} 
\usepackage{aaai25}  
\usepackage{times}  
\usepackage{helvet}  
\usepackage{courier}  
\usepackage[hyphens]{url}  
\usepackage{graphicx} 
\usepackage{natbib}  
\usepackage{caption} 
\usepackage{algorithm}
\usepackage{algorithmic}

\usepackage[pagebackref,breaklinks,colorlinks,bookmarks,citecolor=mycitecolor]{hyperref}
\usepackage{xcolor}
\definecolor{mycitecolor}{RGB}{0, 0, 255} 
\usepackage{titletoc}

\usepackage{multirow}
\usepackage{bbding}
\usepackage{booktabs}
\usepackage{amsmath}
\usepackage{color}
\usepackage{enumitem}
\usepackage{newfloat}
\usepackage{listings}
\DeclareCaptionStyle{ruled}{labelfont=normalfont,labelsep=colon,strut=off} 
\floatstyle{ruled}
\newfloat{listing}{tb}{lst}{}
\floatname{listing}{Listing}
\title{End-to-End Autonomous Driving through V2X Cooperation}
\author{
    Haibao Yu\textsuperscript{\rm 1,2}\equalcontrib,
    Wenxian Yang\textsuperscript{\rm 2}\equalcontrib, 
    Jiaru Zhong\textsuperscript{\rm 2,3}\equalcontrib, 
    Zhenwei Yang\textsuperscript{\rm 2,4}, \\
    Siqi Fan\textsuperscript{\rm 2},
    Ping Luo\textsuperscript{\rm 1},
    Zaiqing Nie\textsuperscript{\rm 2}\thanks{Corresponding author.}
}
\affiliations{
    \textsuperscript{\rm 1}The University of Hong Kong,
    \textsuperscript{\rm 2}AIR, Tsinghua University \\
    \textsuperscript{\rm 3}Beijing Institute of Technology,
    \textsuperscript{\rm 4}University of Science and Technology Beijing

    yuhaibao94@gmail.com, zaiqing@air.tsinghua.edu.cn
}

\begin{document}

\maketitle

\input{sections/0.abstract}
\input{sections/1.introduction}
\input{sections/3.method}
\input{sections/4.experiments}
\input{sections/5.conclusion}

\newpage

\input{sections/6.acknowledgment}

\begin{links}
    \link{Code}{https://github.com/AIR-THU/UniV2X}
\end{links}

\bibliography{reference}

\input{sections/7.appendix}

\end{document}

%% file: sections/0.abstract.tex
\begin{abstract}
  Cooperatively utilizing both ego-vehicle and infrastructure sensor data via V2X communication has emerged as a promising approach for advanced autonomous driving. However, current research mainly focuses on improving individual modules, rather than taking end-to-end learning to optimize final planning performance, resulting in underutilized data potential. In this paper, we introduce UniV2X, a pioneering cooperative autonomous driving framework that seamlessly integrates all key driving modules across diverse views into a unified network. We propose a sparse-dense hybrid data transmission and fusion mechanism for effective vehicle-infrastructure cooperation, offering three advantages: 1) Effective for simultaneously enhancing agent perception, online mapping, and occupancy prediction, ultimately improving planning performance. 2) Transmission-friendly for practical and limited communication conditions. 3) Reliable data fusion with interpretability of this hybrid data. We implement UniV2X, as well as reproducing several benchmark methods, on the challenging DAIR-V2X, the real-world cooperative driving dataset. Experimental results demonstrate the effectiveness of UniV2X in significantly enhancing planning performance, as well as all intermediate output performance. The project is available at \href{https://github.com/AIR-THU/UniV2X}{https://github.com/AIR-THU/UniV2X}.
\end{abstract}

%% file: sections/1.introduction.tex
\section{Introduction}    
\begin{table*}[ht]
\caption{
Comparison with the existing methods for cooperative autonomous driving. ``AgentP" denotes dynamic object perception.``Map" denotes online mapping. ``Occ" denotes occupancy prediction. ``-" denotes that the information is not verified.}
\label{tab: dataset comparison}
\resizebox{\textwidth}{!}{
\begin{tabular}{l|c|cccc|cccc|c}
\hline
\hline
\multirow{2}*{Approach} & \multirow{2}*{Sensor Data} & \multicolumn{4}{c|}{Transmission} & \multicolumn{4}{c|}{Task} & \multirow{2}*{End-to-End} \\
 & & Type & Effective & Friendly & Reliable  & AgentP &  Map & Occ & Plan/Control & \\
\hline
\hline
V2VNet~\cite{wang2020v2vnet} & Point Cloud & BEV Feature & - & Medium & No &  \Checkmark &  & & & No \\
\hline
CoCa3D~\cite{hu2023collaboration}  & Image & BEV Feature & - & Medium & No &  \Checkmark &  & & & No \\
\hline
CoBEVT~\cite{xu2022cobevt} & Image & BEV Feature & - & Medium & No & \Checkmark & \Checkmark & & & No \\
\hline
PP-VIC~\cite{yu2023v2x}  & Point Cloud & Detected Result & - & High & Yes  &  \Checkmark &  & & & No \\
\hline
DeepA~\cite{wang2023deepaccident}  & Image &  BEV Feature & - & Medium & No &  \Checkmark & \Checkmark & & & No \\
\hline
TransIFF~\cite{chen2023transiff} & Point Cloud & Instance Feature & - & High & Yes & \Checkmark  & & & & No \\
\hline
Where2Comm~\cite{hu2022where2comm} & Point Cloud & Instance Feature & - & High & Yes & \Checkmark  & & & & No \\
\hline
CSA~\cite{valiente2019controlling}  & Image & Raw Image & Yes & Low & Yes & & & & \Checkmark & Non-Explicit \\
\hline
CooperNaut~\cite{cui2022coopernaut}  & Point Cloud & BEV Feature & Yes & Medium & No & & & & \Checkmark & Non-Explicit \\
\hline\hline
\textbf{UniV2X (Ours)} & Image & Hybrid Feature &  Yes &  High  &  Yes  &  \Checkmark & \Checkmark & \Checkmark & \Checkmark & Explicit \\
\hline
\hline
\end{tabular}
}
\end{table*}

\begin{figure}[t]
    \centering
    \includegraphics[width=1.0\columnwidth]{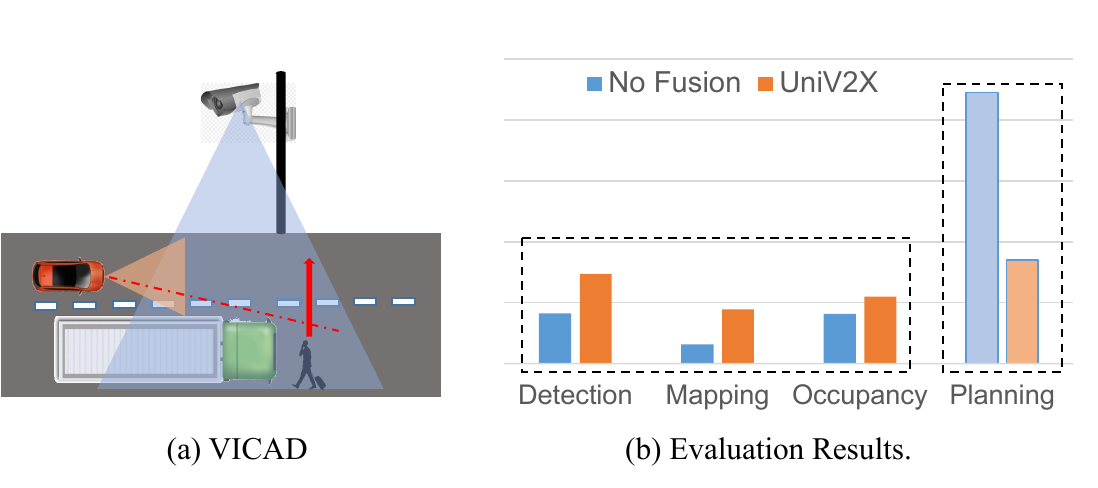}
    \captionsetup{font={scriptsize}}
    \caption{
    (a) VICAD: Infrastructure sensor installed highly has a broad perception field~\cite{yu2022dair,yu2023v2x,yang2023bevheight}, which can supplement the blind and long-range spots of single vehicle.
    (b) Performance Enhancement: Compared with No Fusion solution, UniV2X achieves significant gains in various tasks, such as detection (+13\%), mapping (+11.4\%), occupancy prediction (+5.7\%), and collision rate (-0.5\%).
    }
    \label{fig: vicad and performance comparison}
\end{figure}

Despite significant progress achieved through the integration of deep learning, single-vehicle autonomous driving still faces great safety challenges due to limited perceptual range and inadequate information, especially for vehicles relying on cost-effective cameras. Leveraging external sensors, particularly infrastructure sensors with a broader perception field, has shown promising potential for advancing autonomous driving capacities through Vehicle-to-Everything (V2X) communication (see Fig.~\ref{fig: vicad and performance comparison} (a)).
Several research studies have investigated the efficacy of external sensor data in diverse tasks such as detection \cite{wang2020v2vnet,hu2023collaboration,yu2023flow,wang2023umc,qiu2022distributed}, tracking \cite{yu2023v2x}, segmentation \cite{xu2022cobevt}, localization~\cite{jiang2023roadside,dong2023lidar}, and forecasting \cite{yu2023v2x,ruan2023learning,song2024collaborative}. 
However, existing solutions primarily emphasize individual task optimization, neglecting the overall planning enhancement. This creates challenges in comprehensive data exploitation, driven by a misalignment between individual task goals and final planning objectives.
Thus, end-to-end learning exploration, which directly optimizes the final planning output by harnessing both onboard and external sensor data, becomes necessary. 
In this paper, we focus on vehicle-infrastructure cooperative autonomous driving (VICAD).

The VICAD problem can be formulated as a planning-centric optimization with multiple-view sensor inputs under constrained communication bandwidth. Compared with single-vehicle autonomous driving, VICAD poses additional challenges when addressed through end-to-end learning.
Firstly, the transmitted infrastructure data must be \textit{effective}. It should enhance both critical modules and the final planning performance in autonomous driving. These critical modules encompass dynamic obstacle perception, online mapping, and grid occupancy-based general obstacle detection, providing an explicit scene representation crucial for ensuring the safety of autonomous driving.
Second, the data must be \textit{friendly}. Driven by real-time requirements and limited communication conditions, minimizing transmission costs becomes crucial to mitigate communication bandwidth consumption and reduce latency.
Thirdly, the transmitted data must be \textit{reliable}. it should be \textit{reliable}. Vehicles need interpretable information that can be validated and used judiciously to avoid safety issues such as communication attacks or data corruption.
Addressing these challenges necessitates a well-designed solution for data transmission and cross-view data fusion.

Here are a few straightforward attempts to address the cooperative driving problem through end-to-end learning. CSA~\cite{valiente2019controlling} directly shares and feeds raw images received from other vehicles into basic neural networks for control output. CooperNaut~\cite{cui2022coopernaut} shares features derived from point clouds among vehicles and inputs them into a basic CNN network for the final output. However, these existing solutions rely on a vanilla approach, utilizing simple networks to optimize planning and control outputs. This paradigm lacks explicit modules, compromising safety assurance and interpretability. Especially within intricate urban settings, this approach falls short in ensuring the reliability of the driving system. More comparisons are in Table~\ref{tab: dataset comparison}, and related work is discussed in the appendix.

To this end, we introduce UniV2X, an innovative cooperative autonomous driving framework that seamlessly integrates pivotal modules and cross views into a unified network, as depicted in Figure~\ref{fig:univ2x-framework}. Beyond the final planning task, we address three common tasks for scene representation in autonomous driving: 1) agent perception, encompassing 3D object detection, tracking, and motion forecasting for dynamic obstacle perception, 2) road element (especially lane) detection for online mapping, and 3) grid-occupancy prediction for general obstacle perception.
Inspired by UniAD~\cite{hu2023planning}, we adopt a query-based architecture to establish connections across nodes, encompassing internal modules within infrastructure and ego-vehicle systems, as well as cross-view interactions. In transmission and cross-view interaction, we classify agent perception and road element detection as instance-level representation and occupancy prediction as scene-level representation. We transmit agent queries and lane queries for cross-view agent perception interaction and online mapping interaction. We transmit the occupied probability map, recognizing its dense nature at the scene-level occupancy, for cross-view occupancy interaction. This transmission, termed sparse-dense hybrid transmission, balances sparsity and density in spatial and feature dimensions, respectively.
Cross-view data fusion, such as agent fusion, mainly involves temporal and spatial synchronization, cross-view data matching and fusion, data adaptation for planning and intermediate outputs.
The resulting lightweight approach strengthens dynamic object perception, online mapping, and occupancy modules, thereby enhancing planning performance.
Moreover, the interpretability of queries and occupied probability maps at the instance and scene levels, respectively, fortifies the reliability of the VICAD system, bolstering its transmission integrity and fusion safety.

The contributions are summarized as follows:
\begin{itemize}[leftmargin=9pt]
    \item We pioneer a first explicitly end-to-end framework that unifies vital modules within a single model, advancing the landscape of cooperative autonomous driving. Notably, UniV2X is the first end-to-end framework for VICAD.
    \item We design a sparse-dense hybrid transmission and cross-view data interaction approach, aligning with effectiveness, transmission-friendliness, and reliability prerequisites for end-to-end cooperative autonomous driving.
    \item We reproduce several cooperative methods as benchmarks, as well as instantiating the UniV2X on DAIR-V2X~\cite{yu2022dair}. Experimental results underscore the efficacy of our end-to-end paradigm (see Fig.~\ref{fig: vicad and performance comparison} (b)).
\end{itemize}

%% file: sections/3.method.tex
\section{Method}
\begin{figure*}[ht]
    \setlength{\belowcaptionskip}{0pt}
	\centering
	\includegraphics[width=0.9\textwidth]{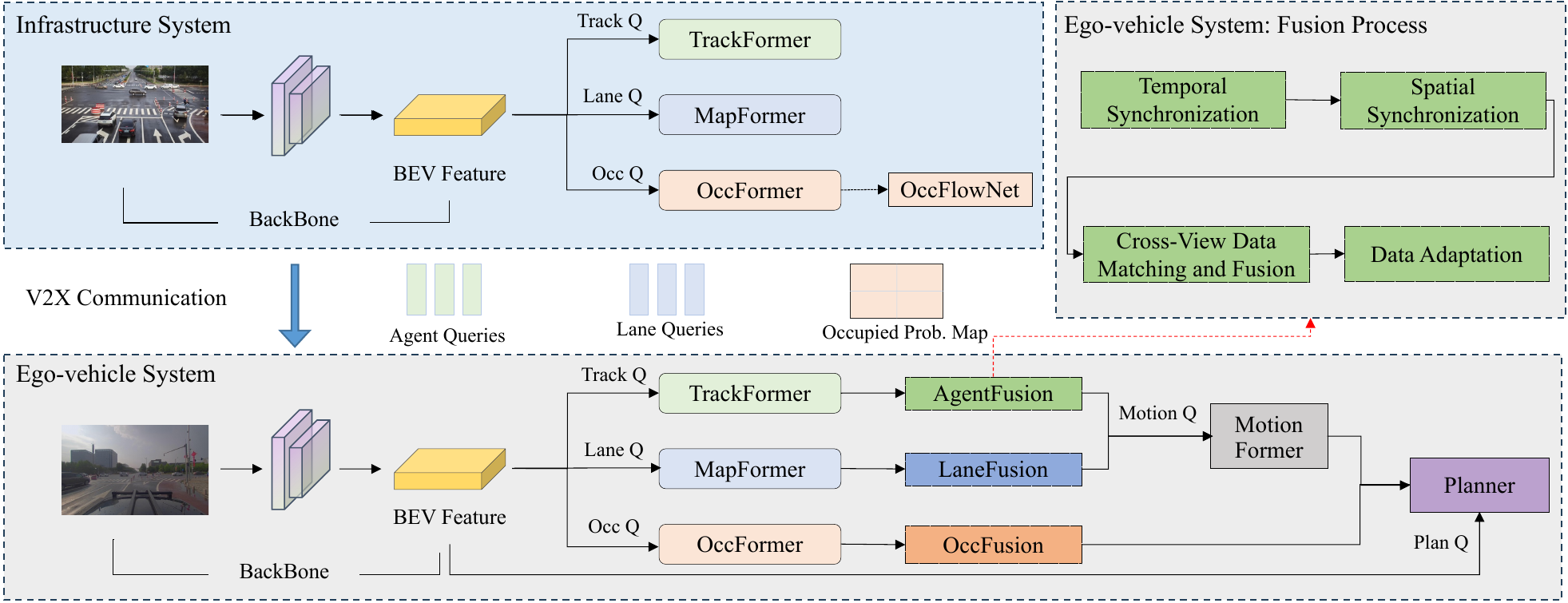}
	\caption{Pipeline of Unified Autonomous Driving through V2X Cooperation (UniV2X). UniV2X aims to connect and jointly optimize all essential modules across diverse views for enhanced planning performance. 
    Cross-view data interaction bolsters pivotal components in autonomous driving like agent perception, online mapping, and occupancy prediction. 
    Additional flow prediction enables minimizing transmission costs for transmitting occupied probability map. 
    Cross-view data fusion involves temporal and spatial synchronization, cross-view data matching and fusion, and data adaptation.}\label{fig:univ2x-framework}
\end{figure*}

In this section, we introduce UniV2X, an all-in-one solution for vehicle and infrastructure cooperative autonomous driving (VICAD), together with the proposed sparse-dense hybrid data transmission and fusion design, as shown in Figure~\ref{fig:univ2x-framework}.
We start by presenting the VICAD problem and introduce the background. Following that, we describe how to generate sparse-dense hybrid data for transmission and cross-view data fusion.
The training process is also outlined.

\subsection{VICAD Problem Formulation}\label{sec: vicad problem}
The VICAD problem is planning-oriented, aiming to improve planning performance by utilizing both infrastructure sensor data and ego-vehicle sensor data through V2X communication. This paper focuses on the images as inputs.
The input of VICAD consists of two parts: 
(a) Ego-vehicle images $\{I_{v}(t)|t\leq t_v\}$ and the relative pose $M_{v}(t_{v})$ at the current vehicle timestamp $t_v$.
(b) Infrastructure images $\{I_{i}(t)|t\leq t_i\}$ and the relative pose $M_{i}(t_{i})$ at the current infrastructure timestamp $t_i$.
Note that in a practical scenario, the timestamp $t_i$ should be earlier than the timestamp $t_v$ due to the communication latency.
The output of VICAD is to predict future coordinates of ego vehicle for time steps $t=t_v+1, \cdots, t_{pred}$.

\subsubsection{Evaluation Metrics.}
We evaluate the planning performance with L2 Error, Collision Rate and Off-Road Rate, and measure transmission cost with Bytes Per Second (BPS), as suggested in~\cite{yu2022dair,yu2023v2x}. We provide detailed explanations for these metrics in the Appendix.

\subsubsection{Challenges.}
Compared with single-vehicle autonomous driving, VICAD presents additional challenges: 
(1) Limited by practical communication conditions, fewer infrastructure data should be transmitted to vehicles to minimize bandwidth usage and reduce latency.
(2) Wireless communication causes latency, potentially leading to temporal misalignment in data fusion.
(3) Potential communication attacks and data corruption can render transmitted data untrustworthy. This highlights the need for interpretable transmitted data.

\subsubsection{Data for Transmission.}
It involves three primary types in V2X cooperation: raw data like raw images, perception outputs like detection results, and intermediate-level data such as Bird's Eye View (BEV) features and queries~\cite{fan2023quest,chen2023transiff,zhong2024leveraging}. Compared to raw data and detection results, intermediate-level data achieves a balance between preserving valuable information and reducing redundant transmission.
To ensure effective, transmission-friendly, and reliable transmitted data, we propose a sparse-dense hybrid transmission mechanism. Queries, as lightweight instance-level features, enhance agent perception and online mapping, as dynamic obstacles and lanes can be treated as instance-level representations. Occupied probability maps, channel-sparse scene-level features, improve occupancy prediction. Compared to less interpretable and high-cost BEV features, occupied probability maps offer pixel-level interpretability and lower transmission costs.

\subsection{Sparse-Dense Hybrid Data Generation}\label{sec: data transmission}
This part illustrates how to generate sparse-dense hybrid data for transmission in infrastructure system. 

BEVFormer\cite{li2022bevformer} is adopted as the backbone to extract image features and transform them into bird's-eye-view (BEV) features $B_{inf}$ with size of (200, 200, 256) by incorporating spatial cross-attention and temporal self-attention. 
TrackFormer is based on DETR~\cite{carion2020end}, which optimizes detection and multi-object tracking together, eliminating the need for non-differentiable post-processing like NMS~\cite{bodla2017soft}. The ultimate filtered output from TrackFormer contains $N_{a}^{inf}$ valid agent queries $\{Q_{A}^{inf}\}$ with a feature dimension of 256 and their corresponding assigned tracking IDs and reference points.
MapFormer is based on Panoptic SegFormer~\cite{li2022panoptic}. We mainly focus on the lane line and crosswalk elements. During transmission, we filter out low-scoring queries using boxes generated from the classification decoder, and exclusively transmit $N_{l}^{inf}$ valid lane queries $\{Q_{L}^{inf}\}$ with a feature dimension of 256, along with their corresponding reference points.
Original OccFormer in UniAD~\cite{hu2023planning} solely considers instance-level occupancy associated with agent queries, predicting multiple steps. However, occupancy serves as a complementary factor to object perception for general obstacle detection, and transmitting multiple probability maps incurs significant transmission costs. To address these challenges, we retain the dense feature obtained through pixel-level attention with a size of (200, 200, 256). Initially, a Multi-layer Perception (MLP) is employed to transform the dense feature into BEV occupied probability map denoted as $P^{inf}$ with a size of (200, 200).
Subsequently, adopting the feature flow prediction approach~\cite{yu2023vehicle,yu2023v2x}, an additional probability flow module is utilized to represent T-step maps via a linear operation as
\begin{equation}\label{eq:feature flow1}
\setlength\abovedisplayskip{0.05cm}
\setlength\belowdisplayskip{0.15cm}
P_{future}(t) = P_0 + t * P_1,
\end{equation}
where $P_0$ signifies the present BEV probability map, and $P_1$ indicates the corresponding BEV probability flow.
Transmitting T-step occupied probability maps requires T*200*200 floats, while UniV2X only requires 2*200*200 floats.

\subsection{Cross-View Data Fusion (Agent Fusion)}\label{sec: cross-view interaction}
In the ego-vehicle system, the BEV features $B_{veh}$ are first extracted from the images captured by onboard sensors. We also adopt TrackFormer, MapFormer, and OccFormer to generate the corresponding agent queries $\{Q_{A}^{veh}\}$, lane queries $\{Q_{L}^{veh}\}$, and the occupied probability map $P^{veh}$. The network for these modules aligns with that of the infrastructure system. 
In this section, we describe how to implement cross-view agent fusion.
Cross-view agent fusion is mainly composed of temporal synchronization for latency compensation, spatial synchronization to unify the cross-view coordinates, data matching and fusing, and data adaptation for planning and intermediate outputs.

\subsubsection{Temporal Synchronization with Flow Prediction.}
The transmission delay in wireless communication, as $t_i$ is earlier than $t_v$, is significant in complex traffic systems, especially for the busy intersection scenario. Due to the movement of dynamic objects, there is a temporal misalignment when fusing data from different sources.
To address that, we incorporate feature prediction into infrastructure agent queries to mitigate latency, following feature flow prediction as~\cite{yu2023vehicle,yu2023v2x}.
Specifically, we input both agent query $Q_A^{inf}$ and query associated in the previous frame into QueryFlowNet, a three-layer Multi-Layer Perceptron (MLP), to generate the agent query flow $Q_{AFlow}^{inf}$. The dimensions of the agent query flow match those of the agent query. Subsequently, a linear operation forecasts future features can be used to mitigate latency $t_v-t_i$, depicted as
\begin{equation}\label{eq:feature flow}
Q_A^{inf}(t_v) = Q_A^{inf}(t_i) + (t_v-t_i) * Q_{AFlow}^{inf}.
\end{equation}
Notably, QueryFlowNet of the Flow Prediction module is not trained in an end-to-end manner in UniV2X. We adopt self-supervised learning following~\cite{yu2023vehicle, yu2023v2x}.

\subsubsection{Spatial Synchronization with Rotation-Aware Query Transformation.}
\begin{figure*}[t]
	\centering
	\includegraphics[width=0.85\textwidth]{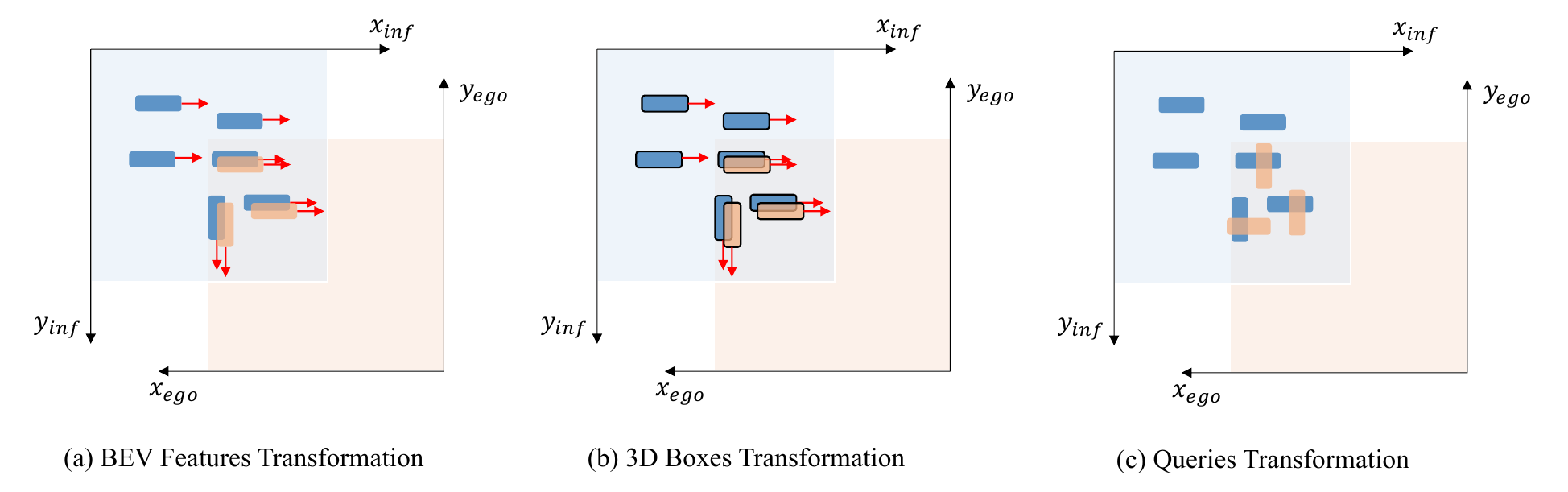}
	\caption{Object orientation is explicitly encoded in BEV feature maps (a) and bounding box (b), while the orientation is implicitly embedded in the feature of queries (c), resulting in the challenge of cross-view rotation alignment in spatial synchronization.}
	\label{fig:rotation-aware}
\end{figure*}

We initially transform the reference points of infrastructure agent queries ${Q_{A}^{inf}}$ from the infrastructure to the ego-vehicle using the relative pose $[R, T]$ between the infrastructure system and ego-vehicle system. Here, the relative pose is generated from the global relative poses of the two systems, with $R$ representing a rotation matrix and $T$ denoting translation. However, each object inherently possesses 3D information about its location, size, and rotation. In the context of a query representing a 3D object, the location is explicitly denoted by reference points, while the rotation is implicitly encoded within the query's feature, as illustrated in Figure~\ref{fig:rotation-aware}.
To address this issue, we propose a solution termed rotation-aware query transformation to achieve spatial synchronization. This involves inputting the infrastructure query, along with its rotation $R$ in the relative pose, into a three-layer MLP to update the feature with rotation awareness, achieving explicit spatial synchronization as
\begin{equation}
\setlength\abovedisplayskip{0.05cm}
\setlength\belowdisplayskip{0.15cm}
\text{spatial\_update}(Q_{A}^{inf}) = \text{MLP}([Q_{A}^{inf}, R]),
\end{equation}
where the rotation matrix $R$ is reshaped into 9 dimensions. Finally, we transform the infrastructure agent query data into the ego-vehicle coordinate system.

\subsubsection{Cross-View Query Matching and Fusion.}
At this stage, cross-view agent queries are temporally and spatially synchronized. To match corresponding queries from different sides, we calculate the Euclidean distance of their reference points and employ the Hungarian method~\cite{kuhn1955hungarian}. For the matched query pairs $Q_{A}^{inf}$ and $Q_{A}^{veh}$, they are fed into a three-layer MLP to generate the cooperated query $Q_{A}$, which is used to update the ego-vehicle agent query $Q_{A}^{veh}$. For the unmatched queries from infrastructure, they are utilized to be added to the ego-vehicle queries. Finally, we assign tracking IDs and filter out cross-view fused queries with low detection confidence, obtaining the final agent queries.

\subsubsection{Ego Identification and Removing.}
This module is used to eliminate the problem of false detection in the ego-vehicle area. From the infrastructure view, the ego vehicle can be perceived either as a distinct obstacle in agent perception or as part of the occupied area in occupancy prediction. Following cross-view data fusion, there is a possibility of generating an obstacle query within the region where the ego vehicle is located, thereby marking the ego-vehicle area as occupied. Such an occurrence can significantly disrupt decision-making processes and ultimately impact decision-making performance.
To mitigate this issue, we define the ego-vehicle area as a rectangle, filter queries within this area, and designate this region as unoccupied. However, this straightforward solution may not consistently perform optimally due to relative position errors between infrastructure and ego vehicle, stemming from positioning and calibration inaccuracies~\cite{yang2023spatio,gu2023feaco}. Further exploration and refinement are essential to contribute to cooperative autonomous driving.

\subsubsection{Decoder Input Augmentation for Intermediate Output.}
Through cross-attention between ultimately fused agent queries and the output from the encoder in the ego-vehicle TrackFormer, we can obtain intermediate outputs for agents, such as 3D detection outputs, to enhance the interpretability of UniV2X. However, the output of the encoder is all generated from ego-vehicle sensor data information, rendering queries from the infrastructure unable to produce corresponding agent outputs. To address this issue, we use synchronized infrastructure queries to enhance ego-vehicle BEV features, the output of the encoder, as:
\begin{equation}
\setlength\abovedisplayskip{0.05cm}
\setlength\belowdisplayskip{0.15cm}
\text{update}(B^{veh}) = B^{veh} + \text{MLP}(\text{synchronized}(Q_{A}^{inf})).
\end{equation}

\subsection{Cross-View Data Fusion (Lane Fusion)}
The LaneFusion module is utilized to fuse lane queries across different sides. In this context, we omit temporal synchronization in lane fusion, as the road lane elements remain unaffected by latency and maintain stability.
Similar to AgentFusion, LaneFusion incorporates spatial synchronization through the rotation-aware query transformation. This process converts infrastructure lane queries, comprising reference points and query features, into the ego-vehicle coordinate system. We then match and fuse the synchronized infrastructure lane queries with ego-vehicle lane queries, as done in AgentFusion. 
To accelerate training, we also choose to directly concatenate synchronized queries with ego-vehicle lane queries.
The synchronized queries are also used for decoder input augmentation.

\subsection{Cross-View Data Fusion (Occupancy Fusion)}
We first generate multiple-step infrastructure occupied probability maps through linear operations, aligning them with ego-vehicle multiple-step occupancy predictions. Leveraging the explicit representation of rotation in the dense probability map, we directly transform the infrastructure occupied probability maps to the ego-vehicle system using the relative pose. Subsequently, we fuse the synchronized occupied probability maps with ego-vehicle occupied probability maps using simple max operations, generating the fused probability map $\hat{P}$.
Grids with a probability exceeding a certain threshold are marked as occupied.

\subsection{Planning Output}
With the fused agent queries, lane queries, and occupancy features, we first generate rough future waypoints by reusing the implementations in UniAD~\cite{hu2023planning}.
MotionFormer is used to generate a set of $N_{a}$ motion queries with a prediction horizon of $t_{pred}$. These queries are created by capturing the interactions among agents, lanes, and goals. Notably, these agent queries encompass the ego-vehicle query, thereby enabling MotionFormer to generate ego-vehicle queries with multimodal intentions. The BEV occupied probability map $\hat{P}$ is utilized to create a binary occupancy map $\hat{O}$.
In the planning phase, the ego-vehicle query obtained from MotionFormer is combined with command embeddings to shape a "plan query". These commands comprise turning left, turning right, and moving forward. This plan query, along with the BEV feature, is input into the decoder to produce future waypoints. 

The final planning trajectory is determined by: 1) adjusting the future waypoints on the roads to ensure adherence to traffic rules and staying within driving areas with the generated lanes and other road elements, and 2) minimizing a cost function to avoid collisions with occupied grids $\hat{O}$. 

%% file: sections/4.experiments.tex
\section{Experiments}
In this section, we implement UniV2X, alongside reproducing various perception, online mapping, and end-to-end methods on DAIR-V2X~\cite{yu2022dair,yu2023v2x}.
More implementation details, ablation studies, visualizations and analysis are provided in the Appendix.
We also conduct UniV2X on more V2X datasets such as V2X-Sim~\cite{li2022v2x}, and present the experiment results in the Appendix.

\begin{table*}[htbp]
\centering
\footnotesize
\renewcommand\arraystretch{1.0}
\setlength\tabcolsep{3.5pt}
\caption{Planning Evaluation Results. We do not report the results at 0.5s and 1.5s because most of the collision rate is zero. \textbf{Although CooperNaut achieves a lower off-road rate, it has a much larger L2 error compared to other methods.} This is because its planning length is relatively conservative, ensuring it is easier to remain within the drivable area over a given period.}
\scalebox{1.0} {
\begin{tabular}{l|cccc|cccc|cccc|c}
\toprule
		\multirow{2}{*}{Method} &
		\multicolumn{4}{c|}{L2 Error ($m$)$\downarrow$} & 
		\multicolumn{4}{c|}{Col. Rate (\%)$\downarrow$} &
        \multicolumn{4}{c|}{Off-Road Rate (\%)$\downarrow$} &
        \multirow{2}{*}{Transm. Cos (BPS)$\downarrow$} \\
		& 2.5$s$ & 3.5$s$ & 4.5$s$ & Avg. & 2.5$s$ & 3.5$s$ & 4.5$s$  & Avg. & 2.5$s$ & 3.5$s$ & 4.5$s$  & Avg. & \\
\midrule
No Fusion & 2.58 & 3.37 & 4.36 & 3.44 & 0.15 & 1.04 & 1.48 & 0.89 & 0.44 & 0.59 & 2.22 & 1.08 & 0\\
Vanilla & 2.33 & 3.69 & 5.12 & 3.71 & 0.59 & 2.07 & 3.70 & 2.12 & 0.15 & 1.33 & 4.74 & 2.07 & 8.19$\times 10^7$ \\
BEV Feature Fusion & 2.31 & 3.29 & 4.31 & 3.30 & 0.0 & 1.04 & 1.48 & 0.83 & 0.44 & 0.44 & 1.91 & 0.93 & 8.19$\times 10^7$ \\
CooperNaut~\cite{cui2022coopernaut} & 3.84 & 5.33 & 6.87 & \textcolor{blue}{5.35} & 0.44 & 1.33 & 1.93 & 1.23 & 0.15 & 0.15 & 1.33 & \textcolor{blue}{0.54} & 8.19$\times 10^7$ \\
\hline
\textbf{UniV2X (Ours)} & 2.59 & 3.35 & 4.49 & 3.48 & 0.0 & 0.44 & 0.59 & \textbf{0.34} & 0.74 & 0.74 & 1.19 & \textbf{0.89} & \textbf{8.09$\times 10^5$} \\ 
\bottomrule
\end{tabular}
}
\label{tab: plan results.}
\end{table*}

\subsection{Experiment Settings}
\paragraph{DAIR-V2X Dataset} comprises approximately 100 scenes captured at 28 complex traffic intersections, recorded using both infrastructure and vehicle sensors. Each scene has a duration ranging from 10 to 25 seconds, capturing data at a rate of 10 Hz, and is equipped with a high-definition (HD) map.
This dataset provides a diverse range of driving behaviors, including actions such as moving forward, turning left, and turning right. To align with nuScenes~\cite{caesar2020nuscenes}, we categorize object classes into four categories (car, bicycle, pedestrian, traffic\_cone). 

\paragraph{Implementation.}
We establish the interest range of the ego vehicle as [-50, 50, -50, 50] meters. The ego-vehicle BEV range shares the same area spanning [-50, 50, -50, 50] meters, with each grid measuring 0.25m by 0.25m. The infrastructure BEV range is set as [0, 100, -50, 50] meters, accounting for the camera's forward sensing range and facilitating more effective utilization of infrastructure data.
The experiments are conducted utilizing 8 NVIDIA A100 GPUs. More implementation details are provided in the appendix.  

\paragraph{Baseline Settings.}
No Fusion only utilizes ego-vehicle images as sensor data input, without any infrastructure data input. 
In Vanilla approach, we employ a simple CNN to fuse infrastructure and ego-vehicle BEV features. The fused BEV feature is reshaped into one dimension and subsequently fed into a Multi-Layer Perceptron (MLP) to generate the planning path. 
In BEV Feature Fusion, we use a CNN to fuse two-side BEV features into a new ego-vehicle BEV feature, and send this new feature into UniAD~\cite{hu2023planning}.
CooperNaut~\cite{cui2022coopernaut} originally employs Point Transformer to aggregate cross-view feature by using the sparse characteristics of point clouds. However, the image is a dense representation, and conducting similar sparse operations, as seen with Where2comm~\cite{hu2022where2comm} in Table~\ref{tab: tracking results.}, results in poor performance.
Therefore, we directly transmit dense BEV features and use CNN to fuse features from both sides, with only one frame input each time to achieve better comparison.
Given the significant role of ego status, such as ego-vehicle velocity, in open-loop end-to-end autonomous driving, as illustrated in~\cite{li2023ego}, we remove the ego-vehicle velocity embedding in all baseline settings for a fair comparison. Additionally, we explore the role of ego-vehicle velocity for UniV2X in the appendix.

\subsection{Experiment Results on DAIR-V2X}
\subsubsection{Planning Results.} We report the planning results in Table~\ref{tab: plan results.}. Compared to No Fusion, UniV2X achieves a 61\% reduction in the average collision rate and a 9.3\% reduction in the average off-road rate, as shown in Table~\ref{tab: plan results.}. Notably, as the planning time increases, the performance improvement becomes more pronounced. This result effectively demonstrates that utilizing infrastructure information can enhance autonomous driving performance, particularly for low-cost monocular solutions.
When compared to Vanilla and BEV Feature Fusion methods, UniV2X significantly outperforms them in terms of average collision rate (0.34 \textit{vs} 2.12, 0.34 \textit{vs} 0.83) and average off-road rate (0.89 \textit{vs} 2.07, 0.89 \textit{vs} 0.93).
Even when compared to CooperNaut, UniV2X still achieves a better average collision rate (0.34 \textit{vs} 1.23). However, there is an abnormal phenomenon where CooperNaut exhibits a much larger L2 error than all other methods up to 5.35$m$. This is because its planning length is relatively conservative, resulting in a shorter planning path that more easily stays within the drivable area, thereby achieving a lower off-road rate.
Furthermore, UniV2X requires significantly less transmission cost compared to the baseline solutions ($8.09\times10^5$ vs $8.19\times10^7$), making it far more transmission-efficient and transmission-friendly.

\subsubsection{Agent Perception Results.} 
We employ various fusion strategies on DAIR-V2X, including No Fusion, Early Fusion (fusing raw infrastructure BEV feature), and Late Fusion (fusing infrastructure detection results with Hungarian method\cite{Kuhn2010TheHM}).
Additionally, we reproduce current SOTA cooperative perception methods on DAIR-V2X, namely V2X-ViT~\cite{xu2022v2x}, Where2comm~\cite{hu2022where2comm}, DiscoNet~\cite{li2021learning}, and CoAlign~\cite{lu2023robust}. For a fair comparison, we standardize inputs (image-only) and evaluation settings. All methods, except for CoCa3D~\cite{hu2023collaboration} based on depth estimation, are re-implemented using BEVFormer~\cite{li2022bevformer}.

\begin{table}[htbp]
\centering
\footnotesize
\renewcommand\arraystretch{1.0}
\setlength{\tabcolsep}{3pt}
\caption{Detection and Multi-Object Tracking Evaluation Results.}
\scalebox{0.9} {
\begin{tabular}{c|cccc}
\toprule
Method & mAP $ \uparrow$ & AMOTA $ \uparrow$ & Trans. Cost $\downarrow$ \\
\midrule
No Fusion & 0.165 & 0.163 & 0 \\
Early Fusion  & 0.243 & 0.209 & 8.19$\times 10^7$ \\
Late Fusion  & 0.196 & \textbf{0.263}  & \textbf{6.60$\times 10^2$} \\
\hline
CoAlign  & 0.240 & 0.234 & 8.19$\times 10^7$ \\
CoCa3D  & 0.226 & - & 4.63$\times 10^6$ \\
V2X-ViT & 0.268 & \textbf{0.287} & 2.56$\times 10^6$ \\
Where2comm  & 0.162 & 0.106 & 5.40$\times 10^5$ \\
DiscoNet  & 0.216 & 0.203 & 1.60$\times 10^5$ \\
V2X-ViT+Where2comm  & 0.178 & 0.071 & 7.22$\times 10^4$ \\
\hline
\textbf{UniV2X (Ours)} & \textbf{0.295}~\textcolor{red}{(+0.13)} & \textbf{0.239}~\textcolor{red}{(+0.076)} & \textbf{6.96$\times 10^4$} \\
\bottomrule
\end{tabular}
}
\label{tab: tracking results.}
\end{table}

We present the evaluation results (car class) for detection and tracking in Table~\ref{tab: tracking results.}.
(1) UniV2X demonstrates a notable enhancement of \textbf{+7.6} and \textbf{+3.0} in AMOTA(\%) compared to No Fusion and Early Fusion. 
(2) UniV2X outperforms CoCa3D, Where2comm, and DiscoNet at similar or less transmission cost.
(3) UniV2X achieves inferior tracking performance compared to tracking-by-detection methods with complex association, such as Late Fusion+AB3DMOT (0.239 \textit{vs} 0.263 at AMOTA), but it significantly outperforms this tracking-by-detection solution in detection (0.295 \textit{vs} 0.196). It is important to note that this tracking-by-detection solution is not suitable for end-to-end autonomous driving.
(4) V2X-ViT exhibits better performance than UniV2X at AMOTA (0.287 \textit{vs} 0.239), but it requires much more transmission cost (2.56$\times 10^6$ \textit{vs} 6.94$\times 10^4$). When we further compress the V2X-ViT transmission to a level similar to UniV2X with Where2comm, there is a significant performance drop (from 0.287 to 0.071 at AMOTA). These outcomes underscore the capability of our infrastructure agent queries and agent fusion module in enhancing agent perception ability with light transmission cost.

\subsubsection{Online Mapping Results.} We implement No Fusion, Early Fusion, and CoBEVT\cite{xu2022cobevt} for online mapping on DAIR-V2X. All methods are re-implemented using BEVFormer~\cite{li2022bevformer}. The mapping performance is reported with Segmentation Intersection over Union (IoU) (\%) as the evaluation metric in Table~\ref{tab: mapping results.}.  UniV2X demonstrates notable improvements in lane perception and crossing perception compared No Fusion, Early Fusion and CoBEVT, respectively. Moreover, compared with Early Fusion and CoBEVT, UniV2X requires less than 1/10th of the transmission cost. These results indicate that infrastructure lane queries and cross-view lane fusion are effective in enhancing online mapping ability.
\begin{table}[t]
\centering
\footnotesize
\renewcommand\arraystretch{1.0}
\setlength{\tabcolsep}{0pt}
\caption{Online Mapping Evaluation Results.}
\scalebox{0.90} {
\begin{tabular}{c|cccc}
\toprule
Method & IoU-Lane (\%)$\uparrow$ & IoU-Crosswalk (\%)$\uparrow$ & Trans. Cost (BPS)$\downarrow$ \\
\midrule
No Fusion & 6.4 & 2.7 & 0 \\
Ealry Fusion  & 16.7 & 17.8  & 8.19$\times 10^7$ \\
CoBEVT  & 15.6 & 16.4 & 2.56$\times 10^6$ \\
\hline
\textbf{UniV2X (Ours)}  & \textbf{17.8}~\textcolor{red}{(+11.4)} & \textbf{19.8}~\textcolor{red}{(+17.1)}  & \textbf{1.47$\times 10^5$} \\
\bottomrule
\end{tabular}
}
\label{tab: mapping results.}
\end{table}

\subsubsection{Occupancy Prediction Results.}
Concerning the evaluation of occupancy prediction, as depicted in Table~\ref{tab: occupanc results.}, UniV2X exhibits notably superior performance compared to No Fusion in both near and far regions. Particularly, UniV2X achieves \textbf{+5.7} and \textbf{+13.4} improvement in IoU-n (\%) and IoU-f (\%) respectively. Here, "IoU-n" and "IoU-f" denote evaluation ranges of 30×30m and 50×50m, respectively. These results underscore the effectiveness of our sparse-dense hybrid transmitted data in significantly enhancing occupancy prediction.
\begin{table}[ht]
\centering
\footnotesize
\caption{Occupancy Prediction Evaluation Results.}\label{tab: occupanc results.}
\scalebox{0.95} {
\begin{tabular}{c|cc}
\toprule  
Method & IoU-n (\%)$\uparrow$ & IoU-f (\%)$\uparrow$  \\
\midrule
No Fusion & 16.3 & 13.1 \\
UniV2X & \textbf{22.0} (\textcolor{red}{+5.7}) & \textbf{26.5} (\textcolor{red}{+13.4})  \\ 
\bottomrule
\end{tabular}
}
\end{table}

\subsection{Ablation Study on Reliability}
We evaluate UniV2X under various communication conditions. Here we assess the impact of data transmission corruption. Additionally, we assess the robustness of UniV2X across different communication bandwidths and latencies, as detailed in the Appendix. We specifically utilize agent queries and fusion to illustrate this reliability.

\begin{figure}[t]
    \centering
    \includegraphics[width=0.95\columnwidth]{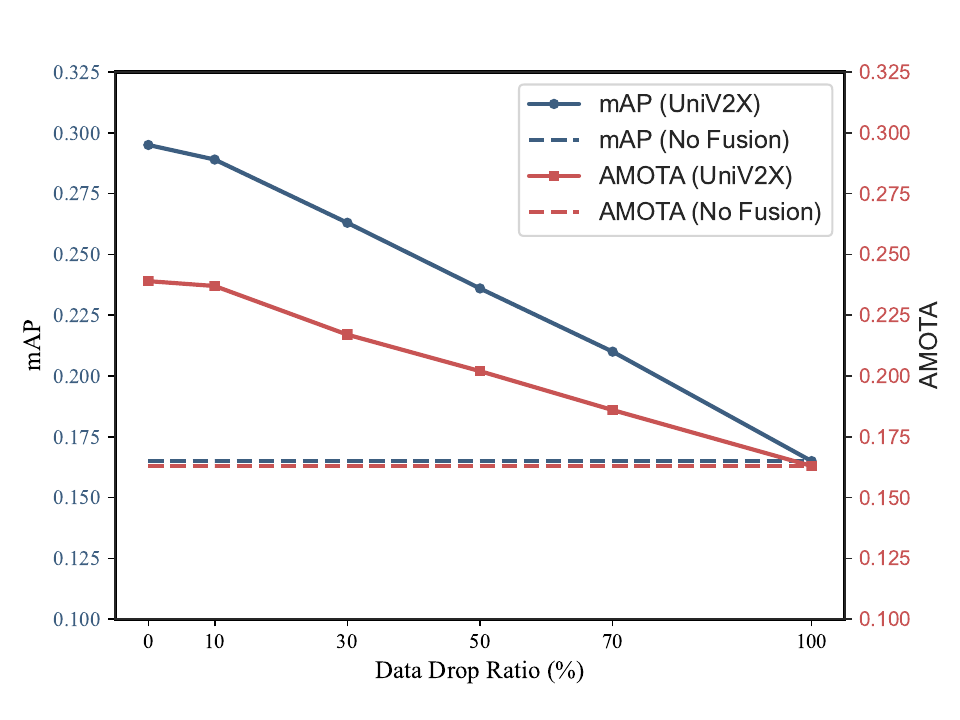}
    \caption{Reliability on Data Corruption.
    }
    \label{fig:data_corruption_agent}
\end{figure}

When assessing UniV2X, our initial step involves randomly discarding 10\%, 30\%, 50\%, 70\%, and 100\% of infrastructure agent queries during transmission to simulate data corruption. Following this, the retained queries are utilized for cross-view query transmission and interaction, and the performance of agent perception, encompassing object detection and tracking, is evaluated accordingly.
The evaluation results depicted in Figure~\ref{fig:data_corruption_agent} unveil a gradual degradation in agent perception as the data corruption ratio increases. When the data corruption ratio reaches 100\%, meaning only ego-vehicle sensor data can be used, the performance of UniV2X becomes comparable to that of the No Fusion model. This decline in performance is anticipated, as data corruption diminishes the complementary information crucial for ego-vehicle autonomous driving. Moreover, even in the absence of certain data due to corruption, UniV2X can maintain a basic level of performance comparable to the No Fusion model. This underscores the reliability of our transmission and cross-view data fusion mechanism.

%% file: sections/5.conclusion.tex
\section{Conclusion}
This paper presents UniV2X, a novel end-to-end framework that integrates crucial tasks from various perspectives into a single network. With a planning-oriented approach, it leverages raw sensor data while ensuring network interpretability for cooperative autonomous driving. Additionally, a sparse-dense hybrid data transmission strategy is devised to harness cross-view data and enhance overall planning performance. This transmission approach is both communication-friendly and reliable, aligning with V2X communication requirements. Empirical results on the DAIR-V2X dataset validate the efficacy of our proposed approach. \\

\noindent\textbf{Limitations and Future Work.}
The framework involves multiple modules and different agent perspectives, resulting in a high degree of complexity. As a result, several interaction fusion modules within the framework remain in preliminary stages. Further refinement is essential for optimizing the internal design of the subsequent framework. 
In this work, we only consider open-loop evaluation for end-to-end autonomous driving. We will conduct more closed-loop experiments to evaluate our UniV2X.

%% file: sections/6.acknowledgment.tex
\section*{Acknowledgment}
This work is supported by the Wuxi Research Institute of Applied Technologies at Tsinghua University under Grant No. 20242001120, as well as the General Research Fund of Hong Kong under Grants No. 17200622 and 17209324.

%% file: sections/7.appendix.tex
\appendix
\clearpage
\newpage

\noindent\textbf{\large{Appendix}}

\startcontents
{
    \hypersetup{linkcolor=black}
    \printcontents{}{1}{}
}

\input{sections/2.related_work}

\section{UniV2X Training}
In our work, we train UniV2X with a four-stage training strategy to ensure comprehensive and stable training.

In the first stage, we pre-train the infrastructure and ego-vehicle systems separately. Specifically, we pre-train the infrastructure system, which includes tasks such as tracking, online mapping, and occupancy prediction, using annotations from the infrastructure view as ground truth. Similarly, we pre-train the ego-vehicle system, which involves tasks like tracking, online mapping, motion prediction, occupancy prediction, and planning, using annotations from the ego-vehicle view as ground truth.

In the second stage, we train the AgentFusion and LaneFusion modules. This stage encompasses tasks such as tracking and online mapping, utilizing cooperative-view annotations as ground truth, with the following combined loss:
\begin{equation} 
\setlength\abovedisplayskip{0.05cm} \setlength\belowdisplayskip{0.15cm} L = L_{\text{track}} + L_{\text{map}}. 
\end{equation}

In the third stage, we train all fusion modules and all modules within the ego-vehicle system, covering all tasks. Cooperative-view annotations are used as ground truth for this stage, with the following combined loss:
\begin{equation} 
\setlength\abovedisplayskip{0.05cm} \setlength\belowdisplayskip{0.15cm} L = L_{\text{track}} + L_{\text{map}} + L_{\text{motion}} + L_{\text{occ}} + L_{\text{plan}}. 
\end{equation}

Furthermore, we use self-supervised learning to train QueryFlowNet and OccFlowNet. This involves constructing infrastructure frame pairs and employing a similarity loss for training, similar to the strategy described in ~\cite{yu2023vehicle}. This training requires no extra annotations.

Note that there is data distribution shift between the generated data from the clips selected from sequences in training stage and generated data from complete sequences in inference stage, especially for the track queries involving complex sequential queries correlation and update in TrackFormer.
We found that if we train the agent fusion with the infrastructure agent queries generated from training clips, there will be unstable performance in inference stage. Thus we save the infrastructure agent queries of training part as the inference way, and train the agent fusion with the saved infrastructure agent queries.
We are also exploring more friendly way to train the agent fusion module.

\section{Evaluation Metric Details}~\label{sec: evaluation metrics.}
In this section, we detail the evaluation metrics used to assess the agent perception, online mapping, occupancy prediction, planning, and transmission cost.

\subsubsection{Object Detection.} We evaluate the detection performance using the mean Average Precision (mAP) metric, which is commonly used in previous works such as~\cite{yu2022dair,caesar2020nuscenes}. The AP is calculated based on the 11-points interpolated precision-recall curve and is defined as follows:
\begin{equation}
\begin{aligned}
    AP & = \frac{1}{11} \sum_{r\in {0.0,...,1.0}} AP_{r} \
        & = \frac{1}{11} \sum_{r\in {0.0,...,1.0}} P_{interp}(r),
\end{aligned}
\end{equation}
where $P_{interp}(r) = \mathop{max}\limits_{\tilde{r}\geq r}p(\tilde{r})$,
and a prediction is considered positive if the Intersection over Union (IoU) is greater than or equal to 0.5. We calculate the AP for each class and then average them to obtain the mAP. In this paper, we only calculate and report Car class.

\subsubsection{Multi-object Tracking.} AMOTA and AMOTP, 3D tracking evaluation metrics \cite{caesar2020nuscenes}, are used to measure the performance of multi-object tracking.

\subsubsection{Online Mapping.}
We categorize the online mapping tasks into two classes: lanes and crosswalks. For each class, we compute the Intersection over Union (\textbf{IoU}) metric, measuring the overlap between the prediction outputs and the ground truths.

\subsubsection{Occupancy Prediction.}
In line with UniAD~\cite{hu2023planning}, we assess the accuracy of predicted occupancy using a methodology similar to that outlined in~\cite{hu2021fiery, zhang2022beverse}. Specifically, the Intersection over Union (\textbf{IoU}) metric quantifies categorical segmentations of the entire scene in an instance-agnostic manner. This metric is computed within Bird's Eye View (BEV) ranges, with evaluations conducted for both near (30m$\times$30m) and far (50m$\times$50m) distances.

\subsubsection{Planning.} We utilize the L2 error, collision rate and off-road rate metrics at different timestamps to assess the planning performance. Compared to existing open-loop evaluations, we have included the off-road rate as an additional metric. This metric is vital for evaluating planning performance, particularly in assessing how well the driving area is maintained and lane violations are minimized.

\subsubsection{Transmission Cost.}
BPS: Byte Per Second (BPS) quantifies the volume of data transmitted from the infrastructure to the ego vehicle per second, accounting for the transmission frequency. In our implementation, a transmission frequency of 2Hz is considered, and we exclude the transmission cost of calibration files and timestamps. For each transmission, the average transmission cost, denoted as $\mathcal{AB}$, is computed. We detail the calculation of the transmission cost for each of the two transmission forms:
\begin{itemize}
    \item Transmitting Queries: Each query is represented as 256-dimensional features in 32-bit float format. Consequently, each query requires eight 256$\times$32-bit floats, equivalent to 1024 Bytes. If ten queries are transmitted per transmission, the $\mathcal{AB}$ of the transmission cost is $1.0\times 10^3$ Bytes. If the transmission frequency is 2Hz, the BPS is $2.0\times 10^3$ Bytes Per Second.
    \item Transmitting Features: Each feature is represented as a tensor. If the size of the feature is $(24, 36, 36)$, and each element is encoded as a 32-bit float, the transmission cost is $24\times 36\times 36\times 4$ Bytes, amounting to $1.2\times 10^5$ Bytes. If the transmission frequency is 2Hz, the BPS is $2.4\times 10^5$ Bytes Per Second.
\end{itemize}

\section{Ablation Study for Fusion Modules}
In this part, we investigate the contributions of each fusion module within UniV2X. 
In evaluating the trained UniV2X, we only fuse agent queries, lane queries, and occupancy probability maps, respectively. We assess each individual fusion module and present the results in Table~\ref{tab: role of agent fusion.}, \ref{tab: role of lane fusion.} and \ref{tab: role of occupancy fusion.}. These tables illustrate that agent fusion, lane fusion, and occupancy fusion each contribute uniquely to agent perception, online mapping, and occupancy prediction, respectively. Additionally, all fusion modules collectively enhance the overall planning performance.
\begin{table*}[htbp]
\centering
\footnotesize
\renewcommand\arraystretch{1.2}
\setlength\tabcolsep{4pt}
\vspace{0pt}
\caption{Evaluation Results of UniV2X with No Fusion and Agent Fusion-Only.}
\scalebox{1.0} {
\begin{tabular}{l|cc|cc}
\toprule
		\multirow{2}{*}{Method} &
		\multicolumn{2}{c|}{Agent Perception} & 
		\multicolumn{2}{c}{Planning}  \\
		& mAP (\%) $\uparrow$ & AMOTA (\%) $\uparrow$ & Col. Rate (\%, Avg.) $\downarrow$ & Off-Road Rate (\%, Avg.) $\downarrow$ \\
\midrule
No Fusion & 16.5 & 16.3 & 0.89 & 1.08 \\
Agent Fusion-Only & 29.5 & 23.9 & 0.59 & 0.79 \\ 
\bottomrule
\end{tabular}
}
\label{tab: role of agent fusion.}
\end{table*}

\begin{table*}[htbp]
\centering
\footnotesize
\renewcommand\arraystretch{1.2}
\setlength\tabcolsep{4pt}
\vspace{0pt}
\caption{Evaluation Results of UniV2X with No Fusion and Lane Fusion-Only.}
\scalebox{1.0} {
\begin{tabular}{l|cc|cc}
\toprule
		\multirow{2}{*}{Method} &
		\multicolumn{2}{c|}{Online Mapping} & 
		\multicolumn{2}{c}{Planning}  \\
		& IoU-Lane (\%) $\uparrow$ & IoU-CrossWalk (\%) $\uparrow$ & Col. Rate (\%, Avg.) $\downarrow$ & Off-Road Rate (\%, Avg.) $\downarrow$ \\
\midrule
No Fusion & 6.4	& 2.7 & 0.89 & 1.08 \\
Lane Fusion-Only & 17.8	& 19.8 & 0.59 & 0.89 \\ 
\bottomrule
\end{tabular}
}
\label{tab: role of lane fusion.}
\end{table*}

\begin{table*}[htbp!]
\centering
\footnotesize
\renewcommand\arraystretch{1.2}
\setlength\tabcolsep{4pt}
\vspace{0pt}
\caption{Evaluation Results of UniV2X with No Fusion and Occ Fusion-Only.}
\scalebox{1.0} {
\begin{tabular}{l|cc|cc}
\toprule
		\multirow{2}{*}{Method} &
		\multicolumn{2}{c|}{Occupancy Prediction} & 
		\multicolumn{2}{c}{Planning}  \\
		& IoU-n (\%) $\uparrow$ & IoU-f (\%) $\uparrow$ & Col. Rate (\%, Avg.) $\downarrow$ & Off-Road Rate (\%, Avg.) $\downarrow$ \\
\midrule
No Fusion & 16.3 & 13.1 & 0.89 & 1.08 \\
Occ. Fusion-Only & 20.3 & 23.5 & 0.39 & 0.98 \\ 
\bottomrule
\end{tabular}
}
\label{tab: role of occupancy fusion.}
\end{table*}

\section{Ablation Study for Flow Prediction}\label{sec: ablation study on latency}
In this section, we explore how our flow prediction can mitigate communication latency. Specifically, we employ Agent Perception to demonstrate this effect.

\subsubsection{Experiment Settings.}
We evaluate UniV2X under two distinct latency conditions: 0$ms$ and 500$ms$. Additionally, to examine the influence of latency on UniV2X's performance, we excluded the prediction module from UniV2X. This version is denoted as UniV2X (without prediction), abbreviated as "UniV2X-O", and was evaluated under both 0 $ms$ and 500 $ms$ latency settings.

\subsubsection{Experiment Results.}
As shown in Table~\ref{tab: latency evaluation results.}, UniV2X-O has a notable decline in perception performance under 500$ms$ latency (a decrease of 3.6\% in mAP and 1.2\% in AMOTA compared to scenarios with no latency). However, with the inclusion of flow prediction, UniV2X demonstrates minimal performance degradation under 500$ms$ latency (only a decrease of 1.2\% in mAP and 0.4\% in AMOTA). These results demonstrate that latency can impact fusion performance and our feature prediction module can mitigate this performance decline.
\begin{table}[htbp]
\centering
\footnotesize
\renewcommand\arraystretch{1.2}
\setlength{\tabcolsep}{3pt}
\caption{Agent Perception Evaluation Results with Latency.}
\scalebox{1.0} {
\begin{tabular}{c|c|cc}
\toprule
Method & Latency (ms) &  mAP (\%)~$\uparrow$ & AMOTA (\%)~$\uparrow$ \\
\midrule
UniV2X-O & 0 & 29.5 & 23.9 \\
UniV2X-O & 500 & 25.9 & 22.7 \\
\hline
UniV2X & 0 & 29.5 & 23.9 \\
UniV2X & 500 & 28.5 & 23.5 \\
\bottomrule
\end{tabular}
}
\label{tab: latency evaluation results.}
\end{table}

\section{Ablation Study for Communication Constraints}\label{sec: ablation study on different communication bandwidth}
In this section, we investigate UniV2X's adaptability to varying communication bandwidths. Specifically, we impose different bandwidth constraints on UniV2X transmission while maximizing the retention of queries and probability maps based on data scores. We present UniV2X's performance across different communication bandwidths in Table~\ref{tab: performance under different communication conditions.}. Across configurations ranging from 0 Mb/s to 1 Mb/s, UniV2X consistently enhances autonomous driving performance, including planning and intermediate outputs. This highlights its ability to adapt to diverse communication bandwidth constraints.

\vspace{5pt}
\begin{table}[htbp]
\centering
\footnotesize
\renewcommand\arraystretch{1.0}
\setlength{\abovecaptionskip}{0.1cm}
\setlength{\tabcolsep}{4.5pt}
\caption{Different Communication Bandwidth Configurations.}
\scalebox{1.0} {
\begin{tabular}{l|ccccc}
\toprule
Bandwidth (Mb/s) & 0.0 & 0.3 & 0.5 & 0.7 & 1.0 \\
\hline
Perception: mAP (\%)  & 16.5 & 21.0 & 23.6 & 26.3 & 29.5 \\
Mapping: Lane (\%)  & 6.4 & 10.9 & 13.2 & 15.0 & 17.7 \\
Occ: IoU-f. (\%) & 13.1 & 20.4 & 23.1 & 25.0 & 26.5 \\
Planning: Col. Rate (\%) & 1.48 & 1.04 & 1.04 & 0.74 & 0.74 \\
\bottomrule
\end{tabular}
}
\label{tab: performance under different communication conditions.}
\end{table}

\section{Experiments on More Dataset}\label{sec: exp on v2x-sim}
In this section, we apply UniV2X to the V2X-Sim dataset~\cite{li2022v2x} to showcase its effectiveness in V2V scenario. Due to the challenges in converting the dataset map to our format, we have not implemented online mapping and lane fusion in our analysis. Consequently, we are unable to report results for the off-road rate metric.

\subsubsection{V2X-Sim Dataset.} 
The V2X-Sim dataset~\cite{li2022v2x} is a synthetic dataset for cooperative autonomous driving, developed through the integration of Carla Simulator~\cite{dosovitskiy2017carla} and SUMO~\cite{lopez2018microscopic} co-simulation. This dataset comprises 100 scenes, and each scene contains synchronized image and point cloud sequences, at a key frame rate of 5Hz, from roadside sensors and sensors mounted on five autonomous driving vehicles.

\subsubsection{Experiment Settings and Results.}
We choose two autonomous vehicles from V2X-Sim and designate one as the ego vehicle to establish V2V scenes. To maintain consistency with the implementation on DAIR-V2X~\cite{yu2022dair}, we equip each autonomous vehicle solely with a front camera sensor. For this experiment, we utilize only one-fifth of the dataset to streamline the results. Considering the absence of evaluation metrics reflecting online mapping's impact on planning, such as lane violation, we exclude the online mapping task from this experiment. We proceed with training and evaluation of UniV2X on V2X-Sim using settings akin to those in DAIR-V2X~\cite{yu2022dair}. Experimental results are detailed in Table~\ref{tab: plan results on v2x-sim.}, ~\ref{tab: agent perception results on v2x-sim.}, and ~\ref{tab: occupancy results on v2x-sim.}.

In comparison to No Fusion, our UniV2X exhibits performance improvements across various aspects, encompassing final planning (with a decrease from 2.0\% to 1.75\% in average Collision Rate) and all intermediate outputs, including agent perception (indicated by an increase from 7.0\% to 7.6\% in mAP) and occupancy prediction (with improvements from 18.9\% to 19.7\% in IoU-far). These experimental results underscore the effectiveness of our UniV2X across a broader spectrum of datasets and V2V scenes.
\begin{table}[htbp]
\centering
\footnotesize
\renewcommand\arraystretch{1.2}
\setlength\tabcolsep{4pt}
\vspace{0pt}
\caption{Planning Evaluation Results on V2X-Sim~\cite{li2022v2x}.}
\scalebox{1.0} {
\begin{tabular}{l|cccc|cccc}
\toprule
		\multirow{2}{*}{Method} &
		\multicolumn{4}{c|}{L2 Error ($m$)$\downarrow$} & 
		\multicolumn{4}{c}{Col. Rate (\%)$\downarrow$}  \\
		& 0.8$s$ & 1.4$s$ & 2.0$s$ & Avg. & 0.8$s$ & 1.4$s$ & 2.0$s$  & Avg. \\
\midrule
No Fusion & 0.98 & 1.79 & 2.65 & 1.81 & 2.0 & 2.25 & 1.75 & 2.0 \\
\textbf{UniV2X} & 0.97 & 1.77 & 2.66 & 1.80 & 1.75 & 2.0 & 1.5 & 1.75 \\ 
\bottomrule
\end{tabular}
}
\label{tab: plan results on v2x-sim.}
\end{table}

\begin{table}[htbp]
\centering
\footnotesize
\renewcommand\arraystretch{1.2}
\setlength\tabcolsep{4pt}
 \caption{Agent Perception Evaluation Results on V2X-Sim~\cite{li2022v2x}.}
\scalebox{1.0} {
\begin{tabular}{l|cc}
\toprule
 method & mAP (\%)$\uparrow$ & AMOTA (\%)$\uparrow$ \\
\midrule
 No Fusion & 7.0 & 0.6 \\
 UniV2X & 7.6 & 1.7 \\
\bottomrule
\end{tabular}
}{
 \label{tab: agent perception results on v2x-sim.}
}
\end{table}

\begin{table}[htbp]
\centering
\footnotesize
\renewcommand\arraystretch{1.2}
\setlength\tabcolsep{4pt}
 \caption{Occupancy Evaluation Results on V2X-Sim~\cite{li2022v2x}.}
\scalebox{1.0} {
\begin{tabular}{l|cc}
\toprule
 method & IoU-n (\%)$\uparrow$ & IoU-f (\%)$\uparrow$ \\
\midrule
 No Fusion & 15.9 & 18.9 \\
 UniV2X & 16.1 & 19.7 \\
\bottomrule
\end{tabular}
}{
 \label{tab: occupancy results on v2x-sim.}
}
\end{table}

\section{Qualitative Visualization}\label{sec: qualitative visualization.}
\begin{figure*}[htbp]
	\centering
	\begin{minipage}[c]{0.7\textwidth}
		\centering
        \includegraphics[width=\textwidth]{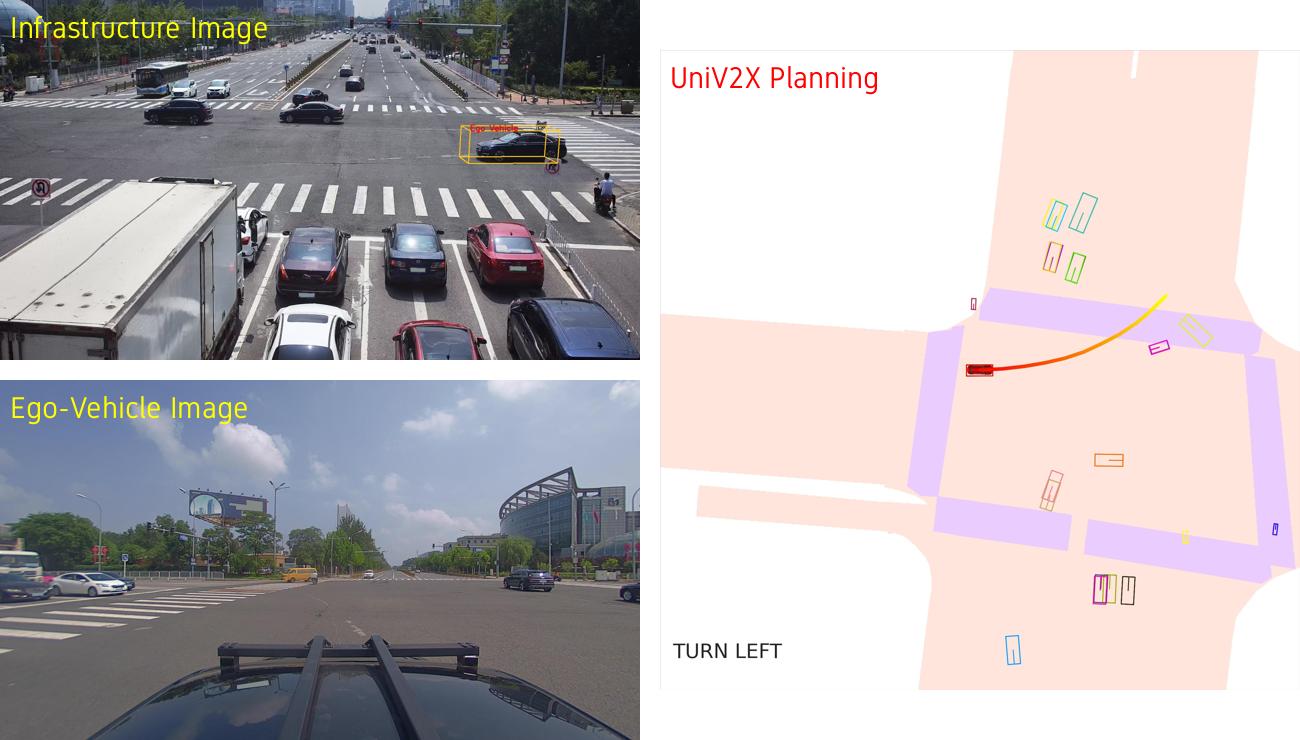}
        \caption{Visualization Example: Turn Left.}
		\label{fig: vis example with turnning left.}
	\end{minipage} \\
	\begin{minipage}[c]{0.7\textwidth}
		\centering
		\includegraphics[width=\textwidth]{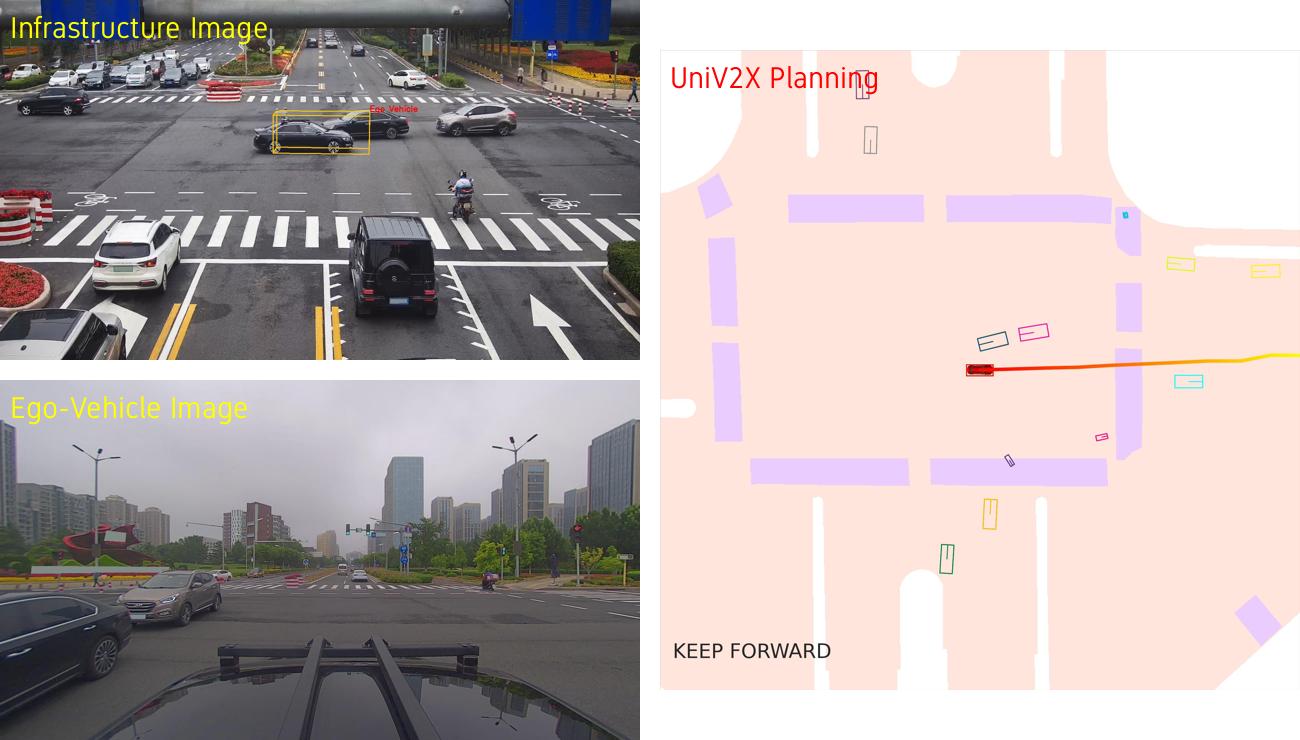}
		\caption{Visualization Example: Keep Forward.}
		\label{fig: vis example with keepping forward.}
	\end{minipage} \\
	\begin{minipage}[c]{0.7\textwidth}
		\centering
		\includegraphics[width=\textwidth]{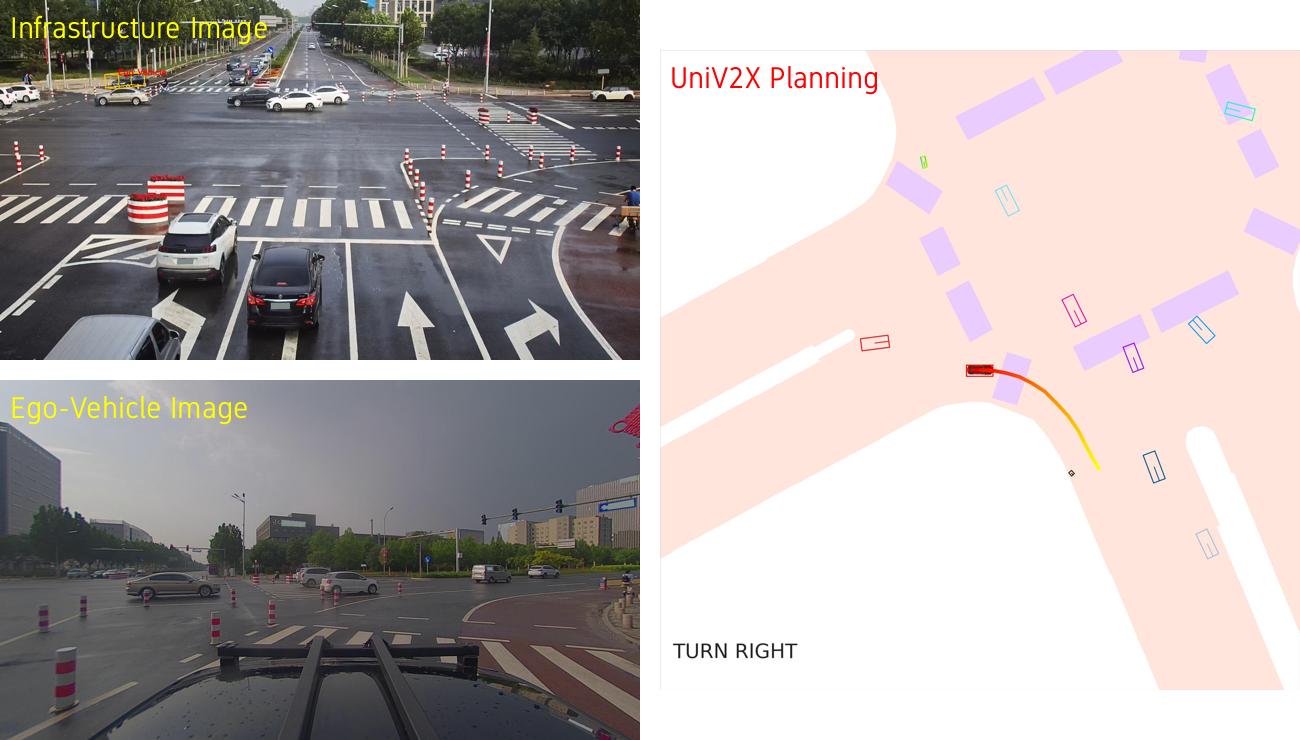}
		\caption{Visualization Example: Turn Right.}
		\label{fig: vis example with turnning right.}
	\end{minipage}
	\caption{Planning Visualization of UniV2X on DAIR-V2X~\cite{yu2022dair}. UniV2X consistently demonstrates the ability to generate high-quality planning outputs for diverse driving scenarios, including left turns, straight keeping, and right turns.}
	\label{fig: vis examples.}
\end{figure*}

In this section, we present visualization examples to demonstrate the efficacy of UniV2X in final planning. The visualization results showcase UniV2X's capability to address diverse driving scenarios, encompassing left turns, straight-ahead navigation, and right turns, as depicted in Figure~\ref{fig: vis examples.}.

%% file: sections/2.related_work.tex
\section{Related Works}
\subsubsection{Cooperative Autonomous Driving.} 
Leveraging V2X communication for cooperative autonomous driving has garnered significant attention. Works like~\cite{wang2020v2vnet, li2021learning,chen2019cooper,yang2023how2comm,hu2023collaboration,lin2024v2vformer,teufel2023collective,yang2024lidar} emphasize transmitting Bird's Eye View (BEV) features extracted from point clouds or images to improve 3D object detection or tracking performance.
\cite{xu2022cobevt} employs sparse transformers to enhance segmentation performance, while\cite{ruan2023learning} focuses on utilizing sequential trajectories for motion forecasting. 
Addressing the latency issue in cooperation detection,~\cite{lei2022latency, yu2023vehicle,wei2023bevflow,yang2023what2comm} utilize historical frames or features.
Additionally,~\cite{hu2022where2comm,fan2023quest,chen2023transiff,zhong2024leveraging} transmit instance-level features or queries to reduce transmission costs and alleviate communication challenges. 
Datasets for cooperative object perception are provided by~\cite{yu2022dair,yu2023v2x,xu2022v2x,li2022v2x,mao2022dolphins,cress2022a9,xu2022opv2v}. ~\cite{yu2023v2x} releases a real-world trajectory dataset generated from infrastructure and vehicle sensor data.
While most of these works concentrate on single tasks, a few also focus on end-to-end output planning or control tasks. For instance,~\cite{valiente2019controlling} employs multi-vehicle images with a simple convolutional network to generate control outputs. Similarly,~\cite{cui2022coopernaut} adopts a simple MLP to learn planning outputs with BEV features from multi-vehicle point clouds. 
However, these solutions lack explicit interpretability modules.
In this paper, we present UniV2X, a comprehensive framework that integrates all essential modules within a single model, utilizing end-to-end learning for optimizing final planning improvement as well as intermediate outputs. 
Recently, several studies~\cite{liu2024towards,chenopencda} have developed benchmarks to evaluate cooperative autonomous driving using closed-loop methodologies. In this work, we focus solely on open-loop evaluation for end-to-end autonomous driving. We plan to conduct closed-loop experiments to assess UniV2X in future work.

\subsubsection{End-to-End Autonomous Driving.}
End-to-end autonomous driving involves the extraction of planning output directly from raw data in a differentiable and learnable manner. Pioneering work, such as~\cite{bojarski2016end}, employs CNNs to generate control outputs from point cloud data. Others, exemplified by~\cite{sadat2020perceive, zeng2020dsdnet}, utilize point clouds and High-definition Maps as inputs. Concurrently, works like~\cite{shao2023safety} leverage multi-modal sensor data as input to generate object density maps for visualization.
For acquiring driving skills,~\cite{codevilla2018end, prakash2021multi, wu2022trajectory} employ imitation learning (IL) to learn from expert demonstrations in an open-loop manner. In contrast,~\cite{liang2018cirl, kendall2019learning, jia2023think} utilize reinforcement learning (RL) to iteratively learn driving skills by interacting with the environment in a closed-loop fashion.
Among related works, UniAD~\cite{hu2023planning} is the first to use queries to connect all essential tasks such as perception, mapping, prediction, and planning. Through unifying these tasks into a single network and employing imitation learning, UniAD achieves remarkable performance on the nuScenes dataset~\cite{caesar2020nuscenes}. However, UniAD solely considers single-vehicle sensor data, while our approach, UniV2X, leverages sensor data from diverse views to enhance the overall driving ability.